\documentclass[sigconf]{acmart}

\pdfoutput=1

\copyrightyear{2018} 
\acmYear{2018} 
\setcopyright{acmcopyright}
\acmConference[ICCAD '18]{IEEE/ACM INTERNATIONAL CONFERENCE ON COMPUTER-AIDED 
	DESIGN}{November 5--8, 2018}{San Diego, CA, USA}
\acmPrice{15.00}
\acmDOI{10.1145/3240765.3240833}

\usepackage{epstopdf}
\usepackage{dblfloatfix}
\usepackage{booktabs}
\usepackage{enumitem}
\usepackage{url}
\usepackage[normalem]{ulem}
\usepackage{subcaption}
\usepackage{color}
\usepackage{adjustbox}
\usepackage{caption} 
\usepackage{framed}

\usepackage{multirow}
\begin{document}

\sloppy



\title{Online Human Activity Recognition using Low-Power Wearable Devices
}

\author{
%
%
Ganapati Bhat$^1$, Ranadeep Deb$^1$,
Vatika Vardhan Chaurasia$^1$, Holly Shill$^2$,
Umit Y. Ogras$^1$ \\
\large{$^1$School of Electrical Computer and Energy Engineering, Arizona State
University, Tempe, AZ \\ $^2$Lonnie and Muhammad Ali Movement Disorder Center, 
Phoenix, AZ}
}
\begin{abstract}
Human activity recognition~(HAR) has attracted significant research interest due to its
applications in health monitoring and patient rehabilitation.  
Recent research on HAR focuses on using smartphones due to their widespread 
use.   
However, this leads to inconvenient use, limited choice of sensors and  inefficient use of resources, 
since smartphones are not designed for HAR.
This paper presents the \textit{first HAR framework 
that can perform both online training and inference}. 
The proposed framework starts with a novel technique 
that generates features using the fast Fourier and discrete wavelet transforms of 
a textile-based stretch sensor and accelerometer.
Using these features, we design an artificial neural network classifier 
which is trained online using the policy gradient algorithm. 
Experiments on a low power IoT device (TI-CC2650 MCU) with nine users 
show 97.7\% accuracy in identifying six activities and their transitions 
with less than 12.5 mW power consumption.
\end{abstract}

\maketitle

\vspace{-3mm}
\section{Introduction}

Advances in wearable electronics has potential to disrupt a wide range of 
health applications~\cite{dinesh2016signal,mosenia2017wearable}.
For example, diagnosis and follow-up for many health problems, such as motion disorders,
depend currently on the behavior observed in a clinical environment.
Specialists  analyze gait and motor functions of patients in a clinic,
and prescribe a therapy accordingly.
As soon as the person leaves the clinic, there is no way to
continuously monitor the patient and report potential
problems~\cite{perez2016dopaminergic,espay2016technology}.
Another high-impact application area is obesity related diseases,
which claim about 2.8 million lives every year~\cite{arif2014better,who_obesity}.
Automated tracking of physical activities of overweight patients, such as walking,
offers tremendous value to health specialists,
since self recording is inconvenient and unreliable.
As a result, human activity recognition (HAR) using low-power wearable devices
can revolutionize health and activity monitoring applications.

There has been growing interest in human activity recognition
with the prevalence of low cost motion sensors and smartphones.
For example, accelerometers in smartphones are used to
recognize activities such as stand, sit, lay down, walking, and 
jogging~\cite{kwapisz2011activity,gyHorbiro2009activity,anguita2013energy}.
This information is used for rehabilitation instruction, fall detection of elderly,
and reminding users to be active~\cite{wang2016comparative,jafari2007physical}.
Furthermore, activity tracking also facilitates physical activity,
which improves the wellness and health
of its users~\cite{bort2014measuring,kirwan2012using,case2015accuracy}.
HAR techniques can be broadly classified based on
when training and inference take place.
Early work collects the sensor data before processing.
Then, both classifier design and inference are performed 
offline~\cite{bao2004activity}.
Hence, they have limited applicability.
More recent work trains a classifier offline,
but processes the sensor data online to infer the 
activity~\cite{anguita2013energy,shoaib2015survey}.
However, to date, there is no technique that can perform
\emph{both online training and inference}.
Online training is crucial, since it needs to adapt to new,
and potentially large number of, users who are not involved in the training process.
To this end, \emph{this paper presents the first HAR technique
that continues to train online to adapt to its user}.

The vast majority, if not all, of recent HAR techniques employ smartphones.
Major motivations behind this choice are their widespread use
and easy access to integrated accelerometer and gyroscope 
sensors~\cite{wang2016comparative}.
We argue that smartphones are not suitable for HAR for three reasons.
First, patients cannot always carry a phone as prescribed by the doctor.
Even when they have the phone, it is not always in the same position (e.g., at hand or in pocket),
which is typically required in these studies~\cite{shoaib2015survey,chen2017performance}.
Second, mobile operating systems are not designed for meeting real-time constraints.
For example, the Parkinson's Disease Dream Challenge~\cite{dream_challenge}
organizers shared raw motion data collected using iPhones in more than 30K experiments.
According to the official spec, the sampling frequency is 100~Hz.
However, the actual sampling rate varies from 89~Hz to 100~Hz,
since the phones continue to perform many unintended tasks during the experiments.
Due to the same reason, the power consumption is in the order of watts
(more than 100$\times$ of our result).
Finally, researchers are limited to sensors integrated in the phones,
which are not specifically designed for human activity recognition.

\begin{figure}[b]
	\centering
	\vspace{-3mm}
	\includegraphics[width=0.99\linewidth]{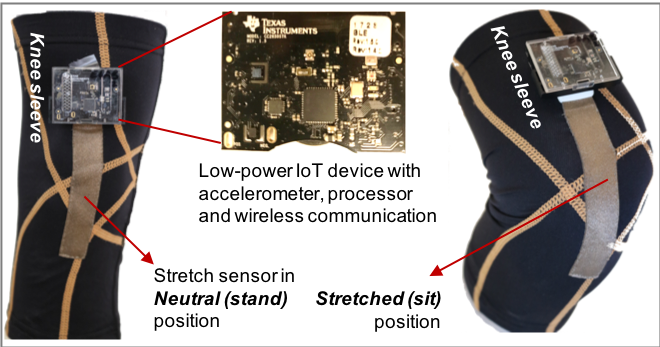}
    \vspace{-2mm}
	\caption{Wearable system setup, sensors and the low-power IoT device~\cite{ticc2650}.
    We knitted the textile-based stretch sensor to a knee sleeve to accurately capture the leg movements.}
	\label{fig:wearable_setup}
\end{figure}

This paper presents an \textit{online} human activity recognition framework
using the wearable system setup shown in Figure~\ref{fig:wearable_setup}.
The proposed solution is the first to perform online training
and leverage textile-based stretch sensors in addition to commonly used accelerometers.
Using the stretch sensor is notable, since it provides low-noise motion data
that enables us to segment the raw data in non-uniform windows
ranging from one to three seconds.
In contrast, prior studies are forced to
divide the sensor data into fixed 
windows~\cite{kwapisz2011activity,arif2014better}
or smoothen noisy accelerometer data over long durations~\cite{chen2017performance} 
(detailed in Section~\ref{sec:related_work}).
After segmenting the stretch and accelerometer data,
we generate features that enable classifying the user activity into
walking, sitting, standing, driving, lying down, jumping, as well as transitions between them.
%
Since the stretch sensor accurately captures the periodicity in the motion,
its fast Fourier transform (FFT) reveals invaluable information 
about the human activity in different frequency bands.
Therefore, we judiciously use the leading coefficients as features in our classification algorithm.
Unlike the stretch sensor, the accelerometer data is notoriously known to be noisy.
Hence, we employ the approximation coefficients of its discrete wavelet transform (DWT)
to capture the behavior as a function of time.
We evaluate the performance of these features for HAR using
commonly used classifiers including artificial neural network, 
random forest,
and k-nearest neighbor~(k-NN).
Among these, we focus on artificial neural network,
since it enables online reinforcement learning
using policy gradient~\cite{sutton1998reinforcement2} with low implementation cost.
Finally, this work is the first to provide a detailed power consumption and performance break-down 
of sensing, processing and communication tasks.
We implement the proposed framework on the TI-CC2650 MCU~\cite{ticc2650},
and present an extensive experimental evaluation using data from nine users and a total of 2614 activity windows.
Our approach provides 97.7\% overall recognition accuracy 
with 27.60~ms processing time, 1.13 mW sensing and 11.24 mW computation power 
consumption.


The major contributions of this work are as follows:
\vspace{-1mm}
\begin{itemize}
	\item A novel technique to segment the sensor data non-uniformly
	as a function of the user motion,
	%
	%
	\item Online inference and training using an NN, 
    and reinforcement learning based on policy gradient,
	\item A low power implementation on a wearable device and
	extensive experimental evaluation of accuracy, performance and power consumption using nine users.
\end{itemize}

The rest of the paper is organized as follows. 
We review the related work in Section~\ref{sec:related_work}.
Then, we present the feature generation and classifier design techniques 
in Section~\ref{sec:feature_classifier}.
Online learning using policy gradient algorithm is detailed in Section~\ref{sec:RL_approach}.
Finally, the experimental results are presented in Section~\ref{sec:experiments}, 
and our conclusions are summarized in Section~\ref{sec:conclusion}.

\vspace{-2mm}
\section{Related Work and Novelty} \label{sec:related_work}
Human activity recognition has been an active area of research due to its
applications in health monitoring, patient rehabilitation and in promoting
physical activity among the general
population~\cite{arif2014better,bort2014measuring, bonomi2009detection}.
Advances in sensor technology have enabled activity recognition to be performed
using body mounted sensors~\cite{preece2009activity}. Typical steps for
activity recognition using sensors include data collection, segmentation,
feature extraction and classification.

HAR studies typically use a fixed window length to infer the activity of a
person~\cite{kwapisz2011activity,arif2014better}. 
For instance, the studies in~\cite{kwapisz2011activity,arif2014better} use 10
second windows to perform activity recognition.
Increasing the window duration improves accuracy~\cite{bonomi2009detection},
since it provides richer data about the underlying activity.
However, transitions between different activities cannot be captured with long windows.
Moreover, fixed window lengths rarely capture the beginning and end of an activity. 
This leads to inaccurate classification as the window can have features of two different
activities~\cite{bonomi2009detection}.
A recent work proposes action segmentation 
using step detection algorithm on the accelerometer data~\cite{chen2017performance}. 
Since the accelerometer data is noisy, 
they need to smoothen the data using a one-second sliding window with 0.5 second overlap. 
Hence, this approach is not practical for low-cost devices with limited memory capacity. 
Furthermore, the authors state that there is a strong need for better segmentation techniques to improve the accuracy of HAR~\cite{chen2017performance}.
To this end, we present a robust segmentation technique 
which produces windows whose sizes vary as a function of the underlying activity.

Most existing studies employ statistical features such as mean, median,
minimum, maximum, and kurtosis to perform
HAR~\cite{pirttikangas2006feature,arif2014better,kwapisz2011activity}.
These features provide useful insight, 
but there is no guarantee that they are representative of all activities. 
Therefore, a number of studies use all the features or choose a
subset of them through feature selection~\cite{pirttikangas2006feature}.
Fast Fourier transform and more recently discrete wavelet transform 
have been employed on accelerometer data. 
For example, the work in~\cite{chen2017performance} computes 
the 5th order DWT of the accelerometer data. 
Eventually, it uses only a few of the coefficients to calculate the wavelet energy in the
0.625 - 2.5 Hz band. 
In contrast, we use only the approximation coefficients of a single level DWT 
with $O(N/2)$ complexity. 
Unlike prior work, we do not use the FFT of the accelerometer data, 
since it entails significant high frequency components without clear implications. 
In contrast, we employ leading FFT coefficients of the stretch sensor data, 
since it gives a very good indication of the underlying activity. 

Early work on HAR used wearable sensors to perform data 
collection while performing various activities~\cite{bao2004activity}. This 
data is then 
processed offline to design the classifier and perform the inference. However, 
offline inference has limited applicability since users do not get any real 
time feedback.
Therefore, recent work on HAR has focused on implementation on 
smartphones~\cite{shoaib2015survey,
	anguita2013energy,bort2014measuring,he2013physical}.
Compared to wearable HAR devices, smartphones have limited choice of sensors 
and high power consumption.
In addition, results on smartphones are harder to reproduce due to the 
variability in different phones,
operating systems and usage patterns~\cite{case2015accuracy,shoaib2015survey}. 

Finally, existing studies on HAR approaches employ
commonly used classifiers, such as k-NN~\cite{friedman2001elements}, support 
vector machines~\cite{friedman2001elements}, 
decision trees~\cite{quinlan2014c4}, and random 
forest~\cite{friedman2001elements}, 
which are 
\emph{trained offline}. 
In strong contrast to these methods, 
the proposed framework is the first to enable \emph{online training}.
We first train an artificial neural network offline to generate an
initial implementation of the HAR system. Then, we use reinforcement learning
at runtime to improve the accuracy of the system. 
This enables our approach to adapt to new users in the field.

\section{\hspace{-2mm}Feature Set and Classifier Design} 
\label{sec:feature_classifier}
\subsection{Goals and Problem Statement} \label{sec:goals_problem}

The goal of the proposed HAR framework is to recognize
the six common daily activities listed in Table~\ref{tab:list_activities} and the transitions between them
\emph{in real-time} with \emph{more than 90\% accuracy} \emph{under mW power range}.
These goals are set to make the proposed system practical for daily use.
The power consumption target enables  day-long operation using
ultrathin lithium polymer cells~\cite{dmi}.


\begin{figure}[t]
	\centering
	\includegraphics[width=0.9\linewidth]{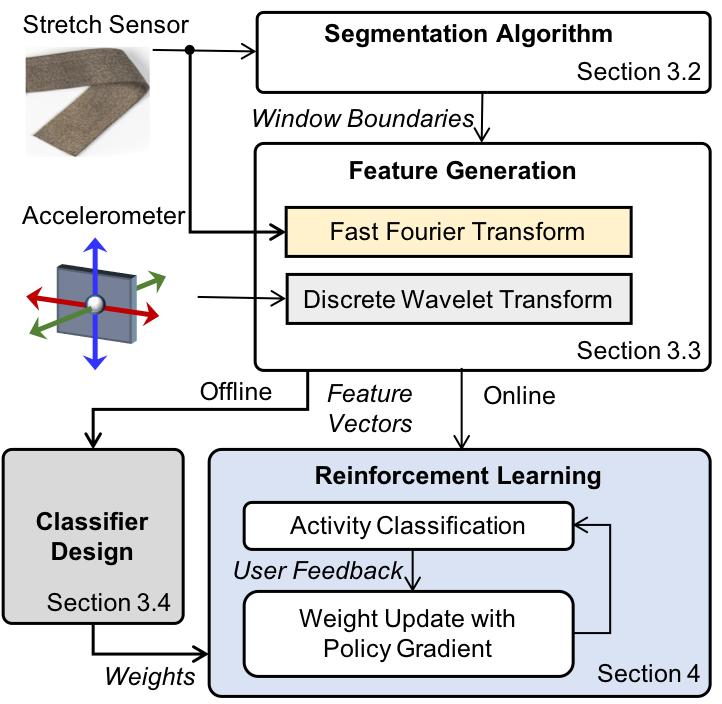}
	\vspace{-1mm}
	\caption{Overview of the proposed human activity recognition framework.}
	\label{fig:overview}
	\vspace{-2mm}
\end{figure}
\begin{table}[t]
	\centering
	\caption{List of activities used in the HAR framework \vspace{-3mm}}
	\label{tab:list_activities}
	\renewcommand{\arraystretch}{0.8}
	\begin{tabular}{@{}lll@{}}
		\toprule 
		\begin{minipage}[t]{0.12\textwidth}
			\begin{itemize}[leftmargin=*]
				\item Drive (D) \end{itemize}
		\end{minipage}
		& \begin{minipage}[t]{0.12\textwidth}\begin{itemize}[leftmargin=*] 
		\item 
				Jump (J) \end{itemize} \end{minipage} &
		\begin{minipage}[t]{0.12\textwidth} \begin{itemize}[leftmargin=*] 
				\item
				Lie Down (L) \end{itemize} \end{minipage}\\ 
		\begin{minipage}[t]{0.12\textwidth} \begin{itemize}[leftmargin=*]\item	
		Sit  
				(S) \end{itemize}  \end{minipage} &
		\begin{minipage}[t]{0.12\textwidth} \begin{itemize}[leftmargin=*]\item 
		Stand 
				(Sd)
		\end{itemize} \end{minipage}
		& \begin{minipage}[t]{0.12\textwidth} 
			\begin{itemize}[leftmargin=*]\item Walk (W)
		\end{itemize} \end{minipage} \\
		\multicolumn{2}{l}{
			\hspace{-1.7mm}\begin{minipage}[t]{0.3\textwidth} 
			\begin{itemize}[leftmargin=*]	
					\item 
					Transition (T) between the activities 
		\end{itemize} \end{minipage}  }		
		\\\bottomrule
		\vspace{-6mm}
	\end{tabular}
\end{table}


The stretch sensor is knitted to a knee sleeve,
and the IoT device with a built-in accelerometer is attached to it,
as shown in Figure~\ref{fig:wearable_setup}.
All the processing outlined in Figure~\ref{fig:overview}
is performed locally on the IoT device.
More specifically, the streaming stretch sensor data is processed
to generate segments ranging from one to three seconds (Section~\ref{sec:segmentation}).
Then, the raw accelerometer and stretch data in each window are processed
to produce the features used by the classifier (Section~\ref{sec:feature}).
Finally, these features are used both for online inference (Section~\ref{sec:classifier})
and reinforcement learning using policy gradient (Section~\ref{sec:RL_approach}).
Since communication energy is significant,
\emph{only the recognized activity and time stamps} are transmitted to
a gateway, such as a phone or PC, using Bluetooth
whenever they are nearby (within ~10m).
The following sections provide a theoretical description of the proposed framework
without tying them to specific parameters values.
These parameters are chosen to enable
a low-overhead implementation using streaming data.
The actual values used in our experiments are summarized in 
Section~\ref{sec:exp_setup}
while describing the experimental setup.

\vspace{-2mm}
\subsection{Sensor Data Segmentation} \label{sec:segmentation}

Activity windows should be sufficiently short to catch 
transitions and fast movements, such as fall and jump. 
However, short windows can also waste computation time and power 
for idle periods, such as sitting. 
Furthermore, a fixed window may contain portions of two different activities, 
since perfect alignment is not possible. 
Hence, activity-based segmentation is necessary to maintain a high accuracy 
with minimum processing time and power consumption. 

\begin{figure}[t]
	\centering
	\includegraphics[width=0.92\linewidth]{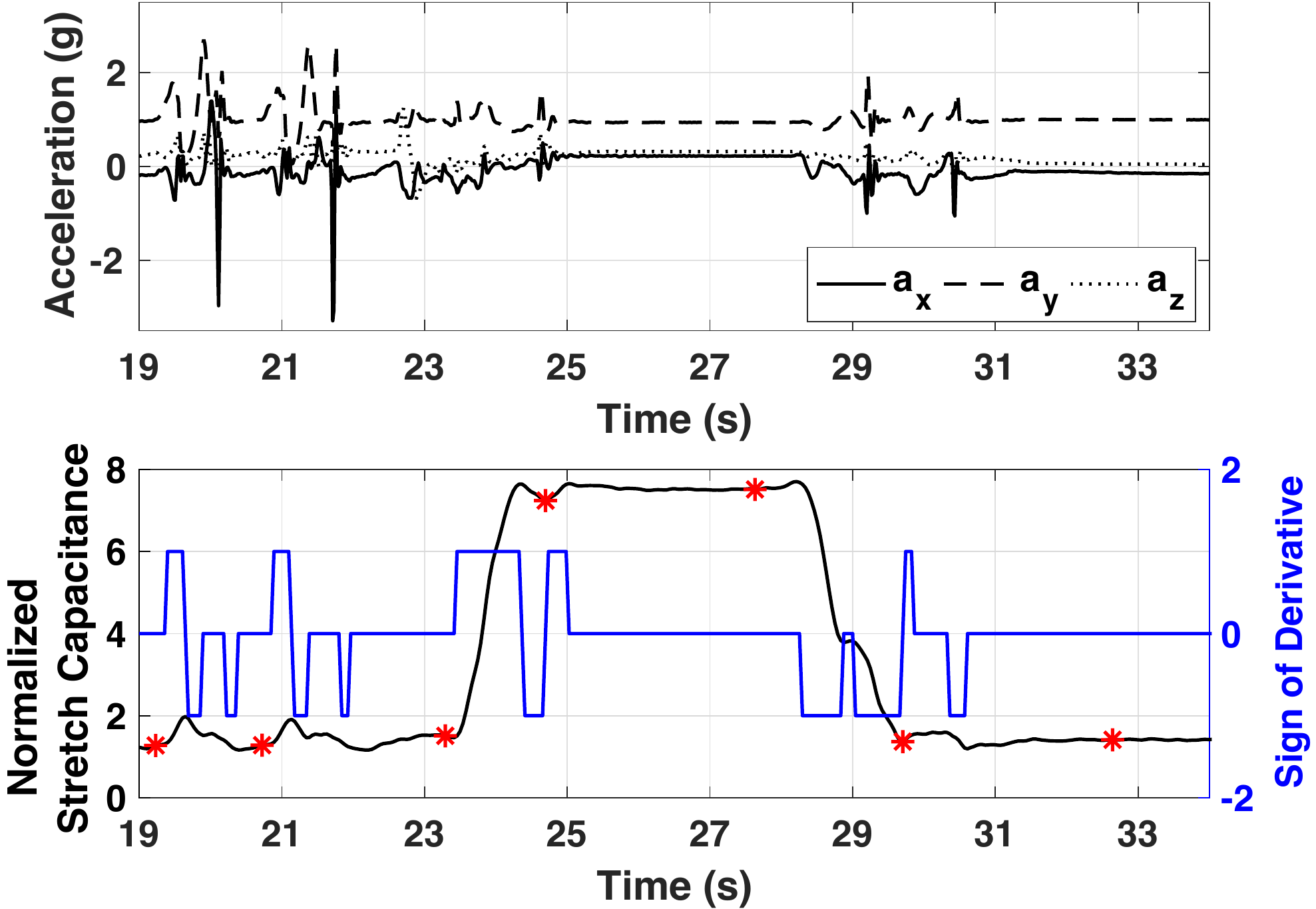}
    \vspace{-3mm}
	\caption{Illustration of the segmentation algorithm.} \label{fig:segmentation}
    \vspace{-5mm}
\end{figure}

To illustrate the proposed segmentation algorithm, 
we start with the snapshot in Figure~\ref{fig:segmentation} from our user studies. 
Both the 3-axis accelerometer and stretch sensor data are preprocessed 
using a moving average filter similar to prior studies. 
The unit of acceleration is already normalized to gravitational acceleration. 
The stretch sensor outputs a capacitance value 
which changes as a function of its state. 
This value ranges from around 390~pF (neutral) 
to close to 500~pF when it is stretched~\cite{o2014stretch}. 
Therefore, we normalize the stretch sensor output 
by subtracting its neutral value and scaling by a constant:
$s(t) = [s_{raw}(t) - min(s_{raw})] / S_{const}$. 
We adopted $S_{const} = 8$ to obtain a comparable range to accelerometer. 
First, we note that the 3-axis accelerometer data exhibits 
significantly larger variations compared to the normalized stretch capacitance. 
Therefore, decisions based on accelerations are prone to false hits~\cite{chen2017performance}.
In contrast, we propose a robust solution which generates 
the segments specified with red \textcolor{red}{$\ast$} markers in Figure~\ref{fig:segmentation}.

The boundaries between different activities can be identified 
by detecting the deviation of the stretch sensor from its neutral value. 
For example, the first segment in Figure~\ref{fig:segmentation} 
corresponds to a step during walk. 
The sensor value starts increasing from a local minima to a peak in the beginning of the step. 
The beginning of the second segment ($t \approx 21$ s) exhibits similar 
behavior, 
since it is another step.
Although the second step is followed by a longer neutral period 
(the user stops and sits to a chair at $t \approx 23$ s),  
the beginning of the next segment is still marked by a rise from a local minima. 
In general, we can observe a distinct minima (fall followed by rise as in walk)
or a flat period followed by rise (as in walk to sit) 
at the boundaries of different activity windows. 
Therefore, the proposed segmentation algorithm 
monitors the derivative of the stretch sensor 
to detect the activity boundaries. 

We employ the 5-point derivative formula given below to 
track the trend of the sensor value:
\vspace{-0.5mm}
\begin{equation}
s'(t) = \frac{s(t-2) - 8s(t-1) + 8s(t+1) - s(t+2)}{12}
\end{equation}
where $s(t)$ and $s'(t)$ are the stretch sensor value 
and its derivative time step $t$, respectively.
When the derivative is positive, we know that the stretch value is \emph{increasing}. 
Similarly, a negative value means a decrease, 
and $s'(t)=0$ implies a flat region. 
Looking at a single data point can catch sudden peaks and lead to false alarms. 
To improve the robustness, one can look at multiple consecutive data points 
before determining the trend. 
In our implementation, we conclude that the \emph{trend} changes 
only if the last three derivatives consistently signal the new trend. 
For example, if the current trend is flat, 
we require that the derivative is positive for three consecutive data points 
to filter glitches in the data point. 
Whenever we detect that the trend changes from flat or decreasing to positive, 
we produce a new segment. 
Finally, we bound the window size from below and above 
to prevent excessively short or long windows. 
We start looking for a new segment, 
only if a minimum duration (one second in this work) 
passes after starting a new window. 
Besides preventing unnecessarily small segments, this approach saves computation time. 
Similarly, a new segment is generated automatically 
after exceeding an upper threshold.  
This choice improves robustness in case a local minima is missed. 
We use $t_{max} = 3$~s as the upper bound, 
since it is long enough to cover all transitions.

\begin{figure}[t]
	\centering
	\includegraphics[width=0.9\linewidth]{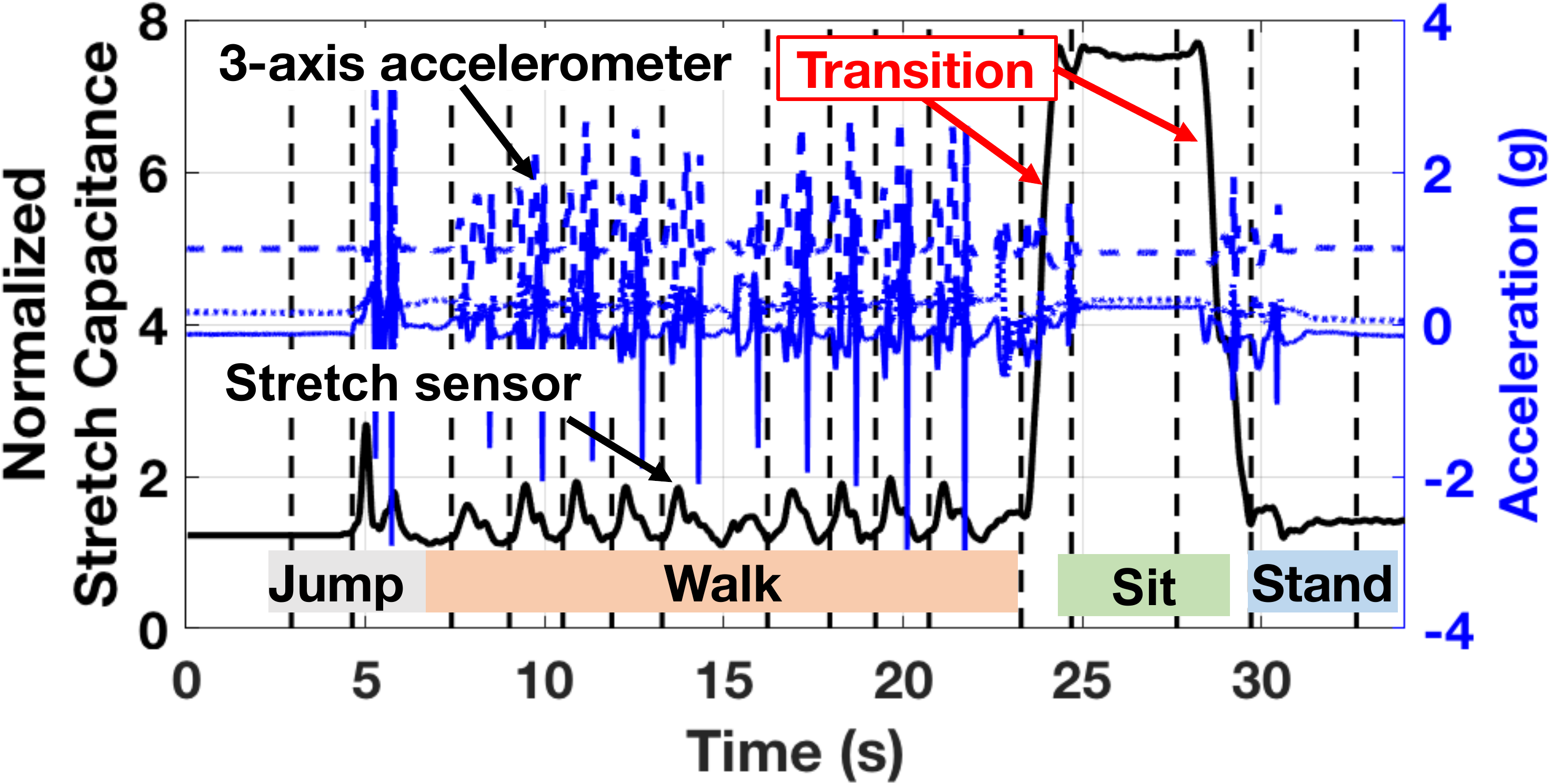}
    \vspace{-4mm}
	\caption{Illustration of the sensor data segmentation.}
	\label{fig:segmented_windows}
	\vspace{-4mm}
\end{figure}

Figure~\ref{fig:segmented_windows} shows the segmented data 
for the complete duration of the illustrative example given in Figure~\ref{fig:segmentation}. 
The proposed approach is able to clearly segment each step of walk. 
Moreover, it is able to capture the transitions from walking to sitting and sitting to
standing very well. This segmentation allows us to extract meaningful features
from the sensor data, as described in the next section.

%

%
\vspace{-2mm}
\subsection{Feature Generation} \label{sec:feature}

To achieve a high classification accuracy,
we need to choose representative features that capture the underlying movements.
We note that human movements typically do not exceed 10-Hz.
Since statistical features, such as mean and variance, are not necessarily representative,
we focus on FFT and DWT coefficients, which have clear frequency interpretations.
Prior studies typically choose the largest transform 
coefficients~\cite{shoaib2015survey}
to preserve the maximum signal power as in compression algorithms.
However, sorting loses the frequency connotation, besides using valuable computational resources.
Instead, we focus on the coefficients in the frequency bins of interest
by preserving the number of data samples in each segment, as described next.

\noindent \textbf{Stretch sensor features:}
The stretch sensor shows a periodic pattern for walking, 
and remains mostly constant during sitting and standing, as shown in Figure~\ref{fig:segmented_windows}.
As the level of activity changes, the segment duration varies in the (1,3] second interval. 
We can preserve 10~Hz sampling rate for the longest duration (3 s during low activity),  
if we maintain $2^5=32$ data samples per segment. 
As the level of activity intensifies, the sampling rate grows to 32~Hz, 
which is sufficient to capture human movements.  
We choose a power of 2, since it enables efficient FFT computation in real-time.
When the segment has more than 32 samples due to larger sensor sampling rate, 
we first sub-sample and smooth the input data as follows:
  \vspace{-1mm}
\begin{equation} \label{Eq:subsample}
s_s[k] = \frac{1}{2S_R}\sum_{i = -S_R}^{S_R} s(tS_R+i), \hspace{7mm} 0 \leq k <
32
\end{equation}
where $S_R = \lfloor N/32 \rfloor$ is the subsampling rate,
and $s_s[k]$ is the sub-sampled and smoothed data point.
When there are less than 32 samples, we simply pad the segment with zeros.

After standardizing the size, we take the FFT of the current window and the previous window.
We use two windows as it allows us to capture any repetitive patterns in the data. 
With 32~Hz sampling rate during high activity regions, we cover $F_s/2=$16 Hz activity per Nyquist theorem. 
We observe that the leading 16 FFT coefficients, which cover the [0-8] Hz frequency range, 
carry most of the signal power in our experimental data.  
Therefore, they are used as features in our classifiers. 
The level of the stretch sensor also gives useful information.
For instance, it can reliably differentiate sit from stand.
Hence, we also add the minimum and maximum value of the stretch sensor to the 
feature set.
%

\noindent \textbf{Accelerometer features:}
Acceleration data contains faster changes compared to the stretch data,
even though the underlying human motion is slow.
Therefore, we sub-sample and smoothen the acceleration to $2^6=64$ points
following the same procedure given in Equation~\ref{Eq:subsample}.
Three axis accelerometers provide acceleration $a_x$, $a_y$ and $a_z$
along $x-$, $y-$ and $z-$axes, respectively.
In addition, we compute the body acceleration 
excluding the effect of gravity $g$ as
$b_{acc} = \sqrt{a_x^2 + a_y^2 + a_z^2} - g$, 
since it carries useful information. 
%
%

Discrete wavelet transform is an effective method to recursively divide
the input signal to approximation $A_i$ and detail $D_i$ coefficients.
One can decompose the input signal to $\log_2 N$ samples where $N$ is the 
number of data points.
After one level of decomposition, $A_1$ coefficients in our data
correspond to 0-32~Hz, while and $D_1$ coefficients cover 32-64~Hz band.
Since the former is more than sufficient to capture acceleration due to human 
activity,
we only compute and preserve $A_1$ coefficients with $O(N/2)$ complexity.
The number of features could be further reduced
by computing the lower level coefficients and preserving largest ones. 
As shown in the performance break-down in Table~\ref{tab:power_perf_energy}, 
using the features in the NN computations takes less time than computing the 
DWT coefficients. 
Moreover, keeping more coefficients and preserving the order maintains the shape of the underlying data. 

\noindent \textbf{Feature Overview:} In summary, we use the following features:

\noindent \emph{Stretch sensor: } We use 16 FFT coefficients,
the minimum and maximum values in each segment. This results in 18 features.

\noindent \emph{Accelerometer: }
We use 32 DWT coefficients for $a_x$, $a_z$ and $b_{acc}$.
In our experiments, we use only the mean value of $a_y$,
since no activity is expected in the lateral direction,
and $b_{acc}$ already captures its effect given the other two directions.
This results in 97 features.

\noindent \emph{General features:} The length of the segment also carries important information,
since the number of data points in each segment is normalized.
Similarly, the activity in the previous window is useful to detect transitions.
Therefore, we also add these two features to obtain a total of 117 features.

%
%
\vspace{-2mm}
\subsection{Supervised Learning for State Classification} \label{sec:classifier}
In the offline phase of our framework, the feature set is assigned 
a label corresponding to the user activity. 
Then, a supervised learning technique takes the labeled data 
to train a classifier which is used at runtime.
Since one of our major goals is online training using reinforcement learning,
we employ a cost-optimized neural network (NN).
\textit{We also compare our solution to most commonly used classifiers by prior work,
and provide brief explanations.}

\noindent\textbf{Support Vector Machine (SVM):}
SVM~\cite{friedman2001elements} finds a hyperplane that can separate the 
feature vectors 
of two output classes. If a separating hyperplane does not exist, SVM maps the 
data into higher dimensions until a separating hyperplane is found. Since SVM 
is a two class classifier, multiple classifiers need to be trained for 
recognizing more than two output classes. Due to this, SVM is not suitable for 
reinforcement learning with multiple 
classes~\cite{lagoudakis2003reinforcement}, which is the case in our HAR 
framework.



\noindent\textbf{Random Forests and Decision Trees:}	Random 
forests~\cite{friedman2001elements} use an
ensemble of tree-structured classifiers, where each tree independently
predicts the output class as a function of the feature vector.
Then, the class which is predicted most often is selected as the final output 
class. C4.5 decision tree~\cite{quinlan2014c4} is another commonly used 
classifier for 
HAR. Instead of using multiple trees, C4.5 uses a single tree.
Random forests typically shows a higher accuracy than decision trees,
since it evaluates multiple decision trees.
Reinforcement learning using random forests has been recently investigated
in~\cite{paul2016reinforced}. As part of the reinforcement learning process, 
additional trees are
constructed and then a subset of trees is chosen to form the new random forest.
This adds additional
processing and memory requirements on the system, making it unsuitable for
implementation on a wearable system with limited memory.

\noindent\textbf{k-Nearest Neighbors~(k-NN):}
k-Nearest Neighbors~\cite{friedman2001elements} is one of the most popular
techniques used by many previous HAR studies. k-NN evaluates the output class
by first calculating k nearest neighbors in the training dataset.
Then, it chooses the class that is most common among the k neighbors and
assigns it as the output class. This requires storing all the training data locally. 
Since storing the training data on a wearable device with limited
memory is not feasible, k-NN is not suitable for online training.

%

\begin{figure}[t]
	\centering
	\vspace{-2mm}
	\includegraphics[width=0.95\linewidth]{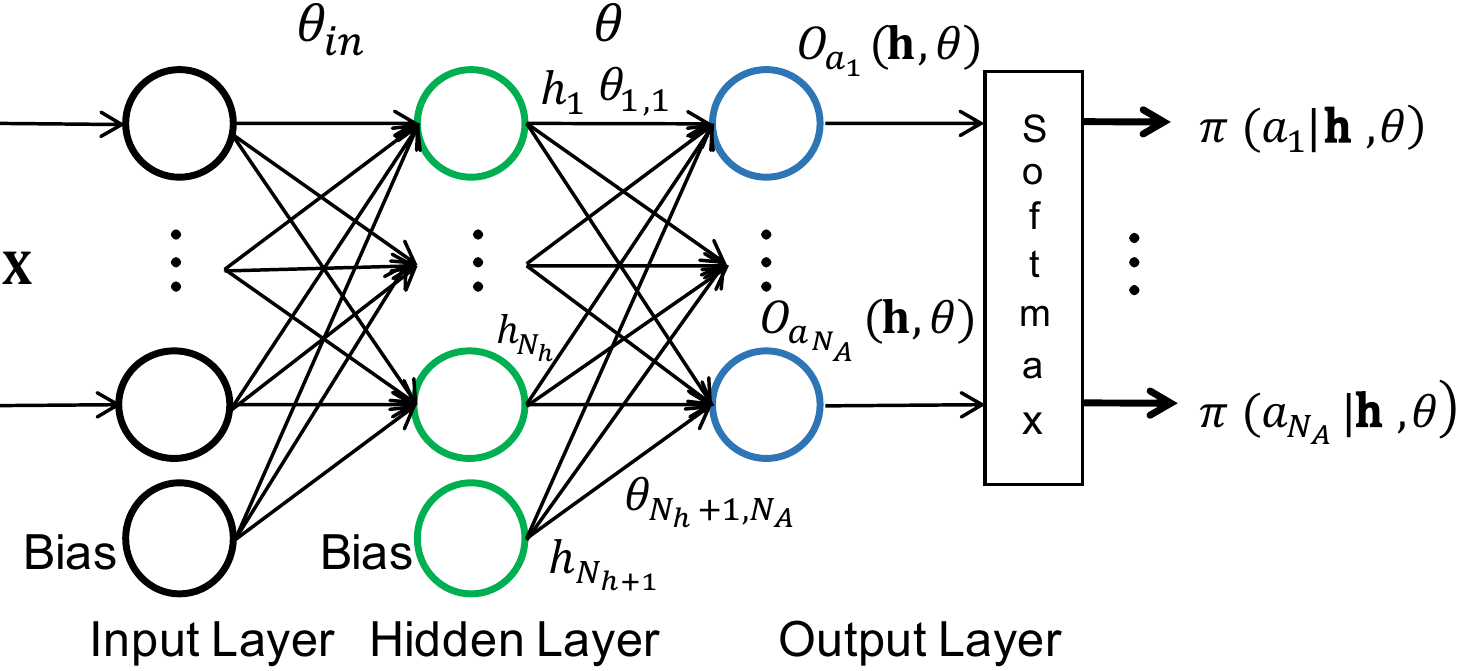}
	\vspace{-3mm}
	\caption{The NN used for activity classifier and reinforcement learning.}
	\label{fig:rl_ann}
	\vspace{-5mm}
\end{figure}

\noindent\textbf{Proposed NN Classifier:}
We use the artificial neural network shown in Figure~\ref{fig:rl_ann} as our 
classifier.
The input layer processes the features denoted by $\mathbf{X}$,
and relay to the hidden layer with the ReLU activation.
It is important to choose an appropriate number of neurons ($N_h$) in the hidden layer
to have a good accuracy, while keeping the computational complexity 
low.
To obtain the best trade-off, we evaluate the recognition accuracy
and memory requirements as a function of neurons, as detailed in 
Section~\ref{sec:ann_training}.

The output layer includes a neuron for each activity
$a_i \in \mathbf{A} = \{D, J, L, S, Sd, W, T\}, 1 \leq i \leq N_A$,
where $N_A$ is the number of activities in set $\mathbf{A}$,
which are listed in Table~\ref{tab:list_activities}.
Output neuron for activity $a_i$ computes 
$O_{a_i}(\mathbf{X},\mathbf{\theta}_{in}, \mathbf{\theta})$
as a function of the input features $\mathbf{X}$ and the weights of the NN.
To facilitate the policy gradient approach described in Section~\ref{sec:RL_approach},
we express the output $O_{a_i}$  in terms of the hidden layer outputs as:
\begin{equation}\label{eq:hidden_out}
\vspace{-1mm}
O_{a_i}(\mathbf{X},\mathbf{\theta}_{in}, \mathbf{\theta}) = 
O_{a_i}(\mathbf{h},\mathbf{\theta}) = \sum_{j=1}^{N_{h} + 1} h_{j} 
\theta_{j,i}, ~~~~~~~1\leq i \leq N_A
\vspace{-1mm}
\end{equation}
where $h_{j}$ is the output of the $j^{\mathrm{th}}$ neuron in the hidden layer,
and $\theta_{j,i}$ is the weight from $j^{\mathrm{th}}$ neuron to output 
activity $a_i$. Note that 
$h_j$ is a function of $\mathbf{X}$ and $\mathbf{\theta}_{in}$.
The summation goes to $N_{h} + 1$,
since there are $N_h$ neurons and one bias term in the hidden layer.

After computing the output functions, we use the softmax activation function
to obtain the probability of each activity:
\begin{equation} \label{Eq:policy}
\vspace{-1mm}
\pi(a_i|\hspace{0.5mm}\mathbf{h}, \mathbf{\theta}) = 
\frac{e^{O_{a_i}(\mathbf{h},\mathbf{\theta})}}
	{\sum_{j=1}^{N_A}{e^{O_{a_j}(\mathbf{h},\mathbf{\theta})}}}, ~~~~~~~1\leq i 
	\leq N_A 
\end{equation}
We express $\pi(a_i|\hspace{0.5mm}\mathbf{h}, \mathbf{\theta})$
as a function of the hidden layer outputs $\mathbf{h}$ instead of the input features,
since our reinforcement learning algorithm will leverage it.
Finally, the activity which has the maximum probability is chosen as the output.

\noindent  \textbf{Implementation cost:} 
Our optimized classifier  requires 264 multiplications for the FFT of stretch data, 
$118N_h + (N_h+1)N_A$ multiplications for the NN and uses only 2~kB memory. 

%

\section{\hspace{-3mm} Online Learning \lowercase{with} Policy Gradient} \label{sec:RL_approach} 

The trained ANN classifier is implemented on the IoT device
to recognize the human activities in real-time.
In addition to online activity recognition, we employ the policy gradient based 
reinforcement learning (RL)
to continue training the classifier in the field.
Online training improves the recognition accuracy for new users
by as much as 33\%, as demonstrated in our user studies.
%
We use the following definitions for the state, action, policy, and the reward.

\noindent\textbf{State:} Stretch sensor and accelerometer readings within a segment
are used as the continuous state space.
We process them as described in Section~\ref{sec:feature} to generate the input 
feature vector $\mathbf{X}$~(Figure~\ref{fig:rl_ann}).

\noindent\textbf{Policy:}
The ANN processes input features as shown in Figure~\ref{fig:rl_ann}
to generate the hidden layer outputs $\mathbf{h} = \{h_j, 1 \leq j \leq N_h+1\}$
and the activity probabilities $\pi(a_i|\mathbf{h}, \mathbf{\theta})$, 
i.e., the policy given in Equation~\ref{Eq:policy}.

\noindent\textbf{Action:}
The activity performed in each sensor data segment
is interpreted as the action in our RL framework. 
It is given by $argmax~\pi(a_i|\mathbf{h}, \mathbf{\theta})$,
i.e., the activity with maximum probability.

%

\noindent\textbf{Reward:}
Online training requires user feedback, which is defined as the reward function.
When no feedback is provided by the user, the weights of the network remain the same.
The user can give feedback upon completion of an activity, such as walking,
which contains multiple segments (i.e., non-uniform action windows).
If the classification in this period is correct, a positive reward (in our implementation $+1$) is given.
Otherwise, the reward is negative ($-1$).
We define the sequence of segments for which a reward is given as an \textit{epoch}. 
The set of epochs in a given training session is called an \textit{episode}
following the RL terminology~\cite{sutton1998reinforcement2}.

\noindent\textbf{Objective:}
The value function for a state is defined as the total reward
that can be earned starting from that state
and following the given policy until the end of an episode.
Our objective is to maximize the total reward $J(\mathbf{\theta})$
as a function of the classifier weights.


\noindent\textbf{Proposed Policy Gradient Update:}
In general, all the weights in the policy network 
can be updated after an epoch~\cite{sutton1998reinforcement2}.
This is useful when we start with an untrained network with random weights.
When a policy network is trained offline as in our example,
its first few layers generate broadly applicable intermediate features~\cite{liang2006fast}.
Consequently, we can update only the weights of the output layer
to take advantage of offline training and minimize the computation cost.
More precisely, we update the weights denoted by $\mathbf{\theta}$ in Figure~\ref{fig:rl_ann}
to tune our optimized ANN to individual users.

Since we use the value function as the objective,
the gradient of $J(\theta)$ is proportional to the gradient of the 
policy~\cite{sutton1998reinforcement2}.
Using this result, the update equation for $\mathbf{\theta}$ is given as:
\begin{equation} \label{eq:weight_update}
\vspace{1mm}
 \mathbf{\theta}_{t+1} \doteq \mathbf{\theta}_t + \alpha r_t
\frac{\nabla_\mathbf{\theta} \pi(a_t|\hspace{0.5mm}\mathbf{h},\mathbf{\theta}_t)}{\pi(a_t|\hspace{0.5mm}\mathbf{h}, \mathbf{\theta}_t)}, 
\hspace{3mm} \alpha : \mathrm{Learning~rate}
\end{equation}
where $\mathbf{\theta}_{t}$ and $\mathbf{\theta}_{t+1}$ are
the current and updated weight matrices, respectively.
Similarly, $a_t$ is the current action at time $t$, $r_t$ is the corresponding reward, 
and $\mathbf{h}$ denotes the hidden layer outputs.
Hence, we need to compute the gradient of the policy to update the weights.  
To facilitate this computation and partial update, we partition the weights into two disjoint sets as
$\mathcal{S}_t$ and $\overline{\mathcal{S}_t}$. 
The weights that connect to the output $O_{a_t}$ corresponding to the current 
action are in $\mathcal{S}_t$. 
The rest of the weights belong to the complementary set $\overline{\mathcal{S}_t}$.
With this definition, we summarize the weight update rule in a theorem 
in order not to disrupt the flow of the paper with derivations. 
Interested readers can go through the proof. 

\vspace{1mm}
\noindent \textbf{Weight Update Theorem:} 
Given the current policy, reward and the learning rate $\alpha$, 
the weights in the output layer of the ANN given in Figure~\ref{fig:rl_ann} 
are updated online as follows:
\begin{equation}\label{eq:weight_update_final}
\theta_{t+1,j,i} \doteq
\begin{cases}
\mathbf{\theta}_{t,j,i} + \alpha r_t (1 - \pi(a_t|\hspace{0.5mm}\mathbf{h},\theta_t)) \cdot 
h_j &
\mathbf{\theta}_{t,j,i} \in  \mathcal{S}_t \\
\mathbf{\theta}_{t,j,i} - \alpha r_t \pi(a_i|\hspace{0.5mm}\mathbf{h},\theta_t)) \cdot 
h_j &
\mathbf{\theta}_{t,j,i} \in  \overline{\mathcal{S}_t}
\end{cases}
\end{equation}

\vspace{2mm}
\noindent \textbf{Proof:}  
The partial derivative of the policy 
$\pi(a_t|\hspace{0.5mm}\mathbf{h}, \mathbf{\theta})$ with respect to the weights $\theta_{j,i}$ 
can be expressed using the chain rule as: 
\begin{equation}\label{eq:partial_derivative}
 \frac{\partial \pi(a_t|\hspace{0.5mm}\mathbf{h},\mathbf{\theta})}{\partial\mathbf{\theta}_{j,i}} =
\frac{\partial \pi(a_t|\hspace{0.5mm}\mathbf{h},\mathbf{\theta})}{\partial 
	O_{a_i}(\mathbf{h},\mathbf{\theta})}
\frac{\partial 
	O_{a_i}(\mathbf{h},\mathbf{\theta})}{\partial\mathbf{\theta}_{j,i}}  
\end{equation}
where $1 \leq j \leq N_h+1$ and $1 \leq i \leq N_A$.
When $\mathbf{\theta}_{t,j,i} \in  \mathcal{S}_t$, action $a_t$ corresponds to output $O_{a_t}(\mathbf{h},\mathbf{\theta})$.
Hence, we can express the first partial derivative using Equation~\ref{Eq:policy} as follows:
\vspace{-1mm}
\begin{align}\label{eq:chain_rule1}
\frac{\partial \pi(a_t|\hspace{0.5mm}\mathbf{h},\mathbf{\theta})}{\partial 
O_{a_t}(\mathbf{h},\mathbf{\theta})} &=
\frac{e^{O_{a_t}(\mathbf{h},\mathbf{\theta})}}
	{\sum_{j=1}^{N_a}{e^{O_{a_j}(\mathbf{h},\mathbf{\theta})}} } -
\frac{\left( e^{O_{a_t}(\mathbf{h},\mathbf{\theta})}\right)^2}
	{\left( \sum_{j=1}^{N_a}{e^{O_{a_j}(\mathbf{h},\mathbf{\theta})}} 
	\right)^2 } 
	\nonumber \\
& = \pi(a_t|\hspace{0.5mm}\mathbf{h},\mathbf{\theta})\big(1 - \pi(a_t|\hspace{0.5mm} \mathbf{h},\mathbf{\theta}) \big)
\end{align}
Otherwise, i.e., $\mathbf{\theta}_{t,j,i} \in  \overline{\mathcal{S}_t}$, 
the derivative is taken with respect to another output. 
Hence, we can find the partial derivative as: 
\begin{align}\label{eq:chain_rule2}
\frac{\partial \pi(a_t|\mathbf{h},\mathbf{\theta})}{\partial 
O_{a_i}(\mathbf{h},\mathbf{\theta})} &=
- \frac{e^{O_{a_t}(\mathbf{h},\mathbf{\theta})} 
e^{O_{a_i}(\mathbf{h},\mathbf{\theta})}}
	{\left( \sum_{j=1}^{N_A}{e^{O_{a_j}(\mathbf{h},\mathbf{\theta})}} 
	\right)^2 
	} 
= -\pi(a_t|\mathbf{h},\mathbf{\theta}) \pi(a_i|\mathbf{h},\mathbf{\theta})
\end{align}
The second partial derivative in Equation~\ref{eq:partial_derivative}, 
$\partial O_{a_i}(\mathbf{h},\mathbf{\theta})/ 
\partial\mathbf{\theta}_{j,i}$,
can be easily computed as $h_j$ using Equation~\ref{eq:hidden_out}.
The weight update is the product of learning rate $\alpha$, reward $r_t$,  $h_j$ 
and the partial derivative of the policy with respect to the output functions.  
For the weights $\mathbf{\theta}_{t,j,i} \in \mathcal{S}_t$, 
we use  the partial derivative in Equation~\ref{eq:chain_rule1}.
For the remaining weights, we use Equation~\ref{eq:chain_rule2}. 
Hence, we obtain the first and second lines in Equation~\ref{eq:weight_update_final}, respectively.
\hfill\textbf{Q.E.D}~~$\square$

In summary, the weights of the output layer are updated online using
Equation~\ref{eq:weight_update_final} after a user feedback. 
Detailed results for the improvement in accuracy
using RL are presented in Section~\ref{sec:rl_results}.

\section{Experimental Evaluation} \label{sec:experiments}
\subsection{Experimental Setup}\label{sec:exp_setup}

\noindent\textbf{Wearable System Setup:} 
The proposed HAR framework is implemented on the TI-CC2650~\cite{ticc2650} IoT 
device, 
which includes a motion processing unit. 
It also integrates a radio that runs Bluetooth Low Energy~(BLE) protocol. 
This device is placed on the ankle, since this allows for a maximum swing in 
the accelerometer~\cite{gyHorbiro2009activity}\footnote{
	We plan to integrate the stretch sensor and the TI-CC2650 into single 
	flexible hybrid electronics device~\cite{gupta2017flexibility}, 
	as shown in Figure~\ref{fig:wearable_setup}, in our future work.}.
The users wear the flexible stretch sensor on the right knee to capture the 
knee movements of the user. 
In our current implementation, the stretch sensor transmits its output to the 
IoT device over BLE to provide flexibility in placement.  
To synchronize the sensors, we record the wall clock time of each sensor at the 
beginning of the experiment.
Then, we compute the offset between the sensors, and use this offset to align 
the sensor readings, as proposed in~\cite{sridhar2016cheepsync}. 
After completing the processing on the IoT device, the recognized activities 
and their time durations 
are transmitted to a host, such as a smartphone, for debugging and offline 
analysis.

\noindent\textbf{Parameter Selection:} 
We use the default sampling frequencies: 100~Hz for the stretch sensor and 
250~Hz for the accelerometer.
Lower sampling frequencies did not produce any significant power savings. 
The raw sensor readings are preprocessed using a moving average filter 
with a window of nine samples.  


\noindent\textbf{User Studies:}
We evaluate the accuracy of the proposed approach using data from nine users,
as summarized in Table~\ref{tab:users_stats}.
The users consist of eight males and one female, 
with ages 20--40 years and heights 160--180 cm.
Data from only five of them are employed during the training phase. 
This data is divided into 80\% training/validation and 20\% test following the 
common practice. 
The rest of the user data is saved for evaluating only the online reinforcement 
learning framework. 
Each user performs the activities listed in Table~\ref{tab:list_activities} 
while wearing the sensors.
For example, the illustration in Figure~\ref{fig:segmented_windows} is from an 
experiment
where the user jumps, takes 10 steps, sits on a chair, and finally stands up. 
The experiments vary from 21 seconds to 6 minutes in length, 
and have different composition of activities. 
We report results from 58 different experiments with a 100 minutes total 
duration, 
as summarized in Table~\ref{tab:users_stats}.  
After each experiment, the segmentation algorithm presented in 
Section~\ref{sec:segmentation} 
is used to identify non-uniform activity windows. 
This results in 2614 unique different segments in our experimental data.
Then, each window is labeled manually through visual inspection by four human 
experts. 
Finally, the labeled data is used for offline training. 
Comparing specific HAR approaches is challenging, 
since data is collected using different platforms, sensors and settings. 
Therefore, we compare our results with all commonly used classifiers in the 
next section.
We also release the labeled experimental data to the public 
on the eLab web 
page\footnote{\url{http://elab.engineering.asu.edu/public-release/}}
to enable other researchers to make comparisons using a common data set.

\begin{table}[h]
	\centering
	\vspace{-4mm}
	\caption{Summary of user studies}
	\vspace{-4mm}
	\label{tab:users_stats}
	\renewcommand{\arraystretch}{0.8}
	\begin{tabular}{@{}cccc@{}}
		\toprule
		Users & Unique Experiments & \begin{tabular}[c]{@{}c@{}}No. of
			Segments\end{tabular} & 		
			Duration (min)\\ \midrule
		9     & 58                 &
		2614                                                             &
		100                                                             \\
		\bottomrule
	\end{tabular}
\vspace{-1mm}
\end{table}

\vspace{-2mm}
\subsection{Training by Supervised Learning}\label{sec:ann_training}
\begin{figure}[h]
	\centering
	\vspace{-0.5mm}
	\includegraphics[width=0.85\linewidth]{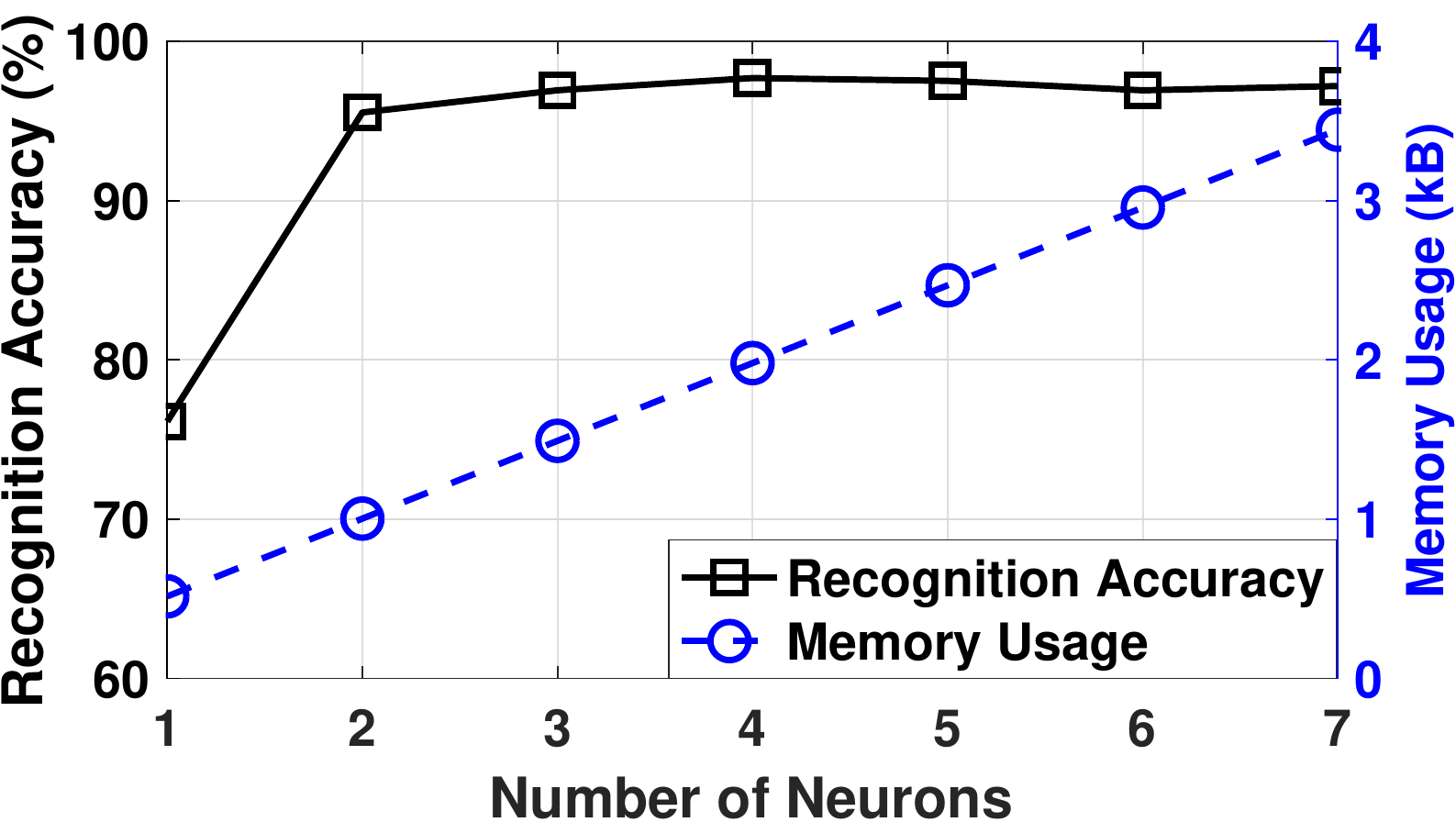}
	\vspace{-3.5mm}
	\caption{Comparison of accuracy with number of neurons}
	\vspace{-5mm}
	\label{fig:neuron_accuracy}
\end{figure}

We use an artificial neural network to perform online activity recognition and training. 
The NN has to be implemented on the wearable device 
with a limited memory (in our case 20kB). 
Therefore, it should have small memory footprint, i.e., number of weights, 
while giving a high recognition accuracy. 
To achieve robust online weight updates during reinforcement learning, 
we first fix the number of hidden layers to one. 
Then, we vary the number of neurons in the hidden layer 
to study the effect on the accuracy and memory requirements.
Specifically, we vary the number of hidden layer neurons from one to seven.
Note that the number of neurons in the output layer remains constant as we do
not change the number of activities being recognized.
Figure~\ref{fig:neuron_accuracy} shows 
the recognition accuracy (left axis) and 
memory requirements (right axis) of the network
as a function of number of neurons in the hidden layer. 
We observe that the accuracy is only about 80\%, 
when a single neuron is used in the hidden layer. 
As we increase the number of neurons, both the memory requirements and accuracy increase. 
The accuracy starts saturating after the third neuron, 
while the number of weights and memory requirements increase.
In fact, the increase in memory requirement is linear,
with an increase of around 500 bytes with every additional neuron in the hidden layer. 
Thus, there is a trade-off between the memory requirements and accuracy. 
In our HAR framework, we choose an NN with four neurons in the
hidden layer as it gives an overall accuracy of about 97.7\% and has a memory
requirement of 2 kB, leaving the rest of the memory for operating system and
other tasks.

%
%
%
%
%
%
%

\vspace{1mm}
\noindent \textbf{5.2.1 Confusion Matrix} 
\vspace{1mm}

\noindent We analyze the accuracy of recognizing each activity in our experiment 
in Table~\ref{tab:confusion_matrix}. 
There is one column and one row corresponding to the activities of interest.  
The numbers on the diagonal show the recognition accuracy for each activity. 
For example, the first row in the first column shows that driving 
is recognized with 99.4\% accuracy. 
According to the first row, 
only 0.6\% of the driving activity windows are classified falsely as ``Transition''. 
To provide also the absolute numbers,  
the number in parenthesis at the end of each row shows the total number of 
activity windows 
with the corresponding label. 
For instance, a total of 155 windows were labeled ``Drive'' according to row 
1.  

We achieve an accuracy greater than 97\%
for five of the seven activities. 
The accuracy is slightly lower for jump because it is more dynamic than all the other activities. Moreover, there is a higher variability in the jump patterns for each user, 
leading to slightly lower accuracy. 
It is also harder to recognize transitions due to the fact that
each transition segment contains features of two activities. 
This can lead to a higher confusion for the NN, but we still achieve more than 
90\% accuracy.
We also note that the loss in accuracy is acceptable for transitions, 
since we can indirectly infer a transition by
looking at the segments before and after the transition.

\begin{table}[t]
	\centering
	
	\caption{Confusion matrix for 5 training users}
	\vspace{-3mm}
	\label{tab:confusion_matrix}
	\renewcommand{\arraystretch}{0.9}
	\setlength\tabcolsep{3.5pt}
	\begin{tabular}{@{}cccccccc@{}} 
		\toprule
		&	Drive & Jump & \begin{tabular}[c]{@{}c@{}}
			Lie \\ Down \end{tabular}& Sit &
		Stand & Walk &
		\begin{tabular}[c]{@{}c@{}}
			Tran-\\ sition \end{tabular} 
		\\
		\midrule
		D~(155) &\textbf{99.4\%}   & 0.00    & 0.00        & 0.00   & 0.00     
		& 0.00    &
		0.6\%         
		\\
		
		J~(181) & 0.00     & \textbf{93.4\% } & 0.00        & 0.00   & 
		1.1\%     & 3.9\%    &
		1.6\%    
		\\
		
		L~(204) & 0.00     & 0.00    & \textbf{100\%}      & 0.00   & 0.00     
		& 0.00    &
		0.00          
		\\
		
		S~(394) & 0.25\%     & 0.25\%    & 0.00        & \textbf{97.7\%} & 
		0.76\%     & 0.00    &
		1.0\%
		\\
		
		Sd~(350) & 0.00     & 0.29\%    & 0.00        & 0.00   & 
		\textbf{98.6\%}   & 
		1.1\%    &
		0.00        
		\\
		
		W~(806) & 0.00     & 0.50\%    & 0.00        & 0.00   & 0.62\%     & 
		\textbf{98.5\%}  &
		0.37\%    
		\\
		
		T~(127) & 0.00     & 3.1\%    & 0.79\%        & 2.4\%   & 0.79\%     & 
		2.4\%    &
		\textbf{90.5\%}   
		\\
		\bottomrule
	\end{tabular}
	\vspace{-3mm}
\end{table}

\vspace{1mm}
\noindent \textbf{5.2.2 Comparison with other classifiers}
\vspace{1mm}

\noindent It is not possible to do a one to one comparison 
with existing approaches because they use different devices, data sets and activities. 
Therefore, we use our data set with the commonly used classifiers
described in Section~\ref{sec:classifier}. 
The results are summarized in Table~\ref{tab:classifier_comparision}. 
Although we use only a single hidden layer and minimize the number of neurons, 
our implementation achieves competitive test and overall accuracy 
compared to the other classifiers. 
We also emphasize that our NN is used for both online
classification and training on the IoT device.

\begin{table}[h]
	\centering
	\caption{Comparison of accuracy for different classifiers}
	\vspace{-3mm}
	\label{tab:classifier_comparision}
	\renewcommand{\arraystretch}{0.8}
	\setlength\tabcolsep{3pt}
	\begin{tabular}{@{}lrrr@{}}
		\toprule
		Classifier    & \begin{tabular}[c]{@{}l@{}}Train
		Acc. (\%)\end{tabular} & \begin{tabular}[c]{@{}l@{}}Test
		Acc. (\%)\end{tabular} & \begin{tabular}[c]{@{}l@{}}Overall
		Acc. (\%)\end{tabular} \\ \midrule
		Random Forest &
		100.00                                                   &
		94.58                                                   &
		98.92                                                      \\
		C4.5          &
		99.09                                                    &
		93.90                                                   &
		98.05                                                      \\
		k-NN          &
		100.00                                                   &
		94.80                                                   &
		98.96                                                      \\
		SVM           &
		97.68                                                    &
		95.03                                                   &
		97.15                                                      \\
		Our NN  &
		98.53                                                    &
		94.36                                                   &
		97.70                                                      \\
		\bottomrule
	\end{tabular}
\vspace{-3mm}
\end{table}

\subsection{Reinforcement Learning with new users} \label{sec:rl_results}
\begin{figure}[t]
	\centering
	\includegraphics[width=1\linewidth]{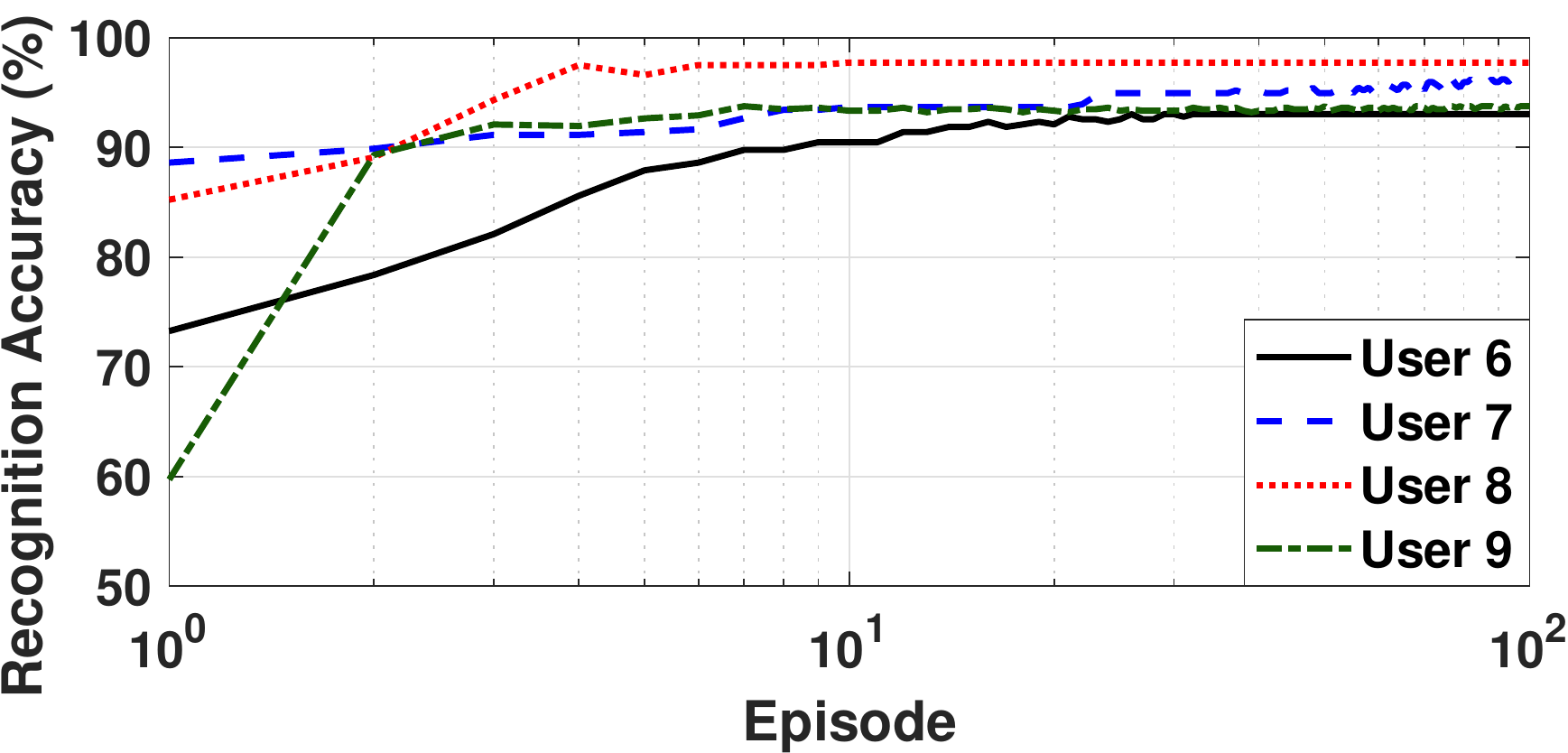}
	\vspace{-5mm}
	\caption{Reinforcement learning results for four new users.}
	\label{fig:rl_new_users}
	\vspace{-3mm}
\end{figure}

The NN obtained in the offline training stage is used to recognize the
activities of four new users that are previously unseen by the network. 
This capability provides a real world evaluation of the approach, since a device
cannot be trained for all possible users. Due to variations in usage patterns,
it is possible that the initial accuracy for a new user is low. Indeed, the
initial accuracy for users 6 and 9 is only about 60 -- 70 \%. Therefore, we use
reinforcement learning using policy gradients to continuously adapt the HAR system to each user. 
Figure~\ref{fig:rl_new_users} shows the improvement achieved using reinforcement learning for four users. 
Each episode in the x-axis corresponds to an iteration of RL using the data set for new users. 
The weights of the NN are updated after each segment as a function of the user 
feedback for a total of 100 episodes. 
Moreover, we run 5 independent runs, each consisting of 100 epochs, 
to obtain  an average accuracy of the NN at each episode. 
We observe consistent improvement in accuracy for all four users. 
The accuracy for users 6 and 9 starts low and increases to about 93\% after 
about 20 episodes. 
User 8 starts with a higher accuracy of about 85\%. 
The accuracy increases quickly to about 98\% after 10 episodes.
In summary, reinforcement improves the accuracy for users not
previously seen by the network. This ensures that the device can adapt to new 
users very easily.

\subsection{Power, Performance and Energy Evaluation}
To fully assess the cost of the proposed HAR framework, 
we present a detailed breakdown of execution time, 
power consumption and energy consumption for each step. 
The first part of HAR involves data acquisition from the sensors and segmentation. 
The segmentation algorithm is continuously running while the data is being acquired. 
Therefore, we include its energy consumption in the sensing block. 
Table~\ref{tab:power_perf_energy} shows the power and energy consumption
for a typical segment of 1.5~s. 
The average power consumption for the data acquisition is 1.13 mW, 
leading to a total energy consumption of 1695 $\mu$J.
If the segments are of a longer duration, the energy consumption for data
sensing increases linearly.
Following the data segmentation, we extract the features and run the classifier. 
The execution time, power and energy for these blocks are shown in
the "Compute" rows in Table~\ref{tab:power_perf_energy}. As expected, the FFT
block has the largest execution time and energy consumption. However, it is
still two orders of magnitude lower than the duration of a typical segment.
Finally, the energy consumption of the BLE communication block is given in the
last row of Table~\ref{tab:power_perf_energy}. Since we transmit the inferred
activity, the energy consumed by the BLE communication is only about 43 $\mu$J.
In summary, with less than 12.5 mW average power consumption, 
our approach enables close to 60-hour uninterrupted operation  
using a 200mAh @ 3.7V battery~\cite{dmi}. Hence, it can enable self-powered 
wearable devices~\cite{bhat2017near} 
that can harvest their own energy~\cite{park2017flexible}.

\begin{table}[h]
	\centering
	\caption{Execution time, power and energy consumption}
	\vspace{-3mm}
	\label{tab:power_perf_energy}
	\renewcommand{\arraystretch}{0.8}
	\setlength\tabcolsep{3pt}
	\begin{tabular}{@{}llrrr@{}}
		\toprule
		& Block            & \begin{tabular}[c]{@{}c@{}}Exe. \\ Time (ms)
		\end{tabular}& \begin{tabular}[c]{@{}c@{}}Average \\ Power (mW)
		\end{tabular}& Energy
		($\mu$J) \\ \midrule
	\multirow{1}{*}{Sense}	& \begin{tabular}[c]{@{}l@{}} Read/Segment\end{tabular}
		 & 1500.00 & 1.13  & 1695.00\\ \midrule
	\multirow{4}{*}{Compute}	& DWT              & 7.90           &
	9.50
		&                   75.05           \\
		& FFT              & 17.20          &     11.80
		&                  202.96            \\
		& NN              & 2.50           &     12.90
		&                    32.25          \\
	& Overall
	 & 27.60 & 11.24 & 310.26 \\ 
	\midrule
	{Comm.} &	BLE & 8.60 & 5.00 & 43.00
		\\ \bottomrule
	\end{tabular}
\end{table}
\vspace{-2mm}
\section{Conclusions} \label{sec:conclusion} 
We presented a HAR framework on a wearable IoT device
using stretch accelerometer sensors.
The first step of our solution is a novel technique to segment the sensor data
non-uniformly as a function of the user motion.
Then, we generate FFT and DWT features using the segmented data.
Finally, these features are used for online inference and training using an ANN.
Our solution is the first to perform online training.
Experiments on TI-CC2650 MCU with nine users 
show 97.7\% accuracy in identifying six activities and their transitions 
with less than 12.5 mW power consumption.

\vspace{-2mm}



\bibliographystyle{abbrv}
\bibliography{references/embedded_refs,references/flexible,references/health_refs}

\end{document}